\documentclass[conference, a4paper]{IEEEtran}
\usepackage{graphicx}
\usepackage{comment}
\usepackage{amsmath,amssymb} 
\usepackage{color}

\usepackage{times}
\usepackage{epsfig}
\usepackage{graphicx}
\usepackage{amsmath}
\usepackage{amssymb}
\usepackage{multirow}
\usepackage{multicol}
\usepackage{xcolor}
\usepackage{transparent}
\usepackage{booktabs}
\usepackage{wrapfig}
\usepackage{float}
\usepackage{afterpage}
\usepackage{hyperref}

\definecolor{8}{rgb}{0.973,0.412,0.42}
\definecolor{7}{rgb}{0.961,0.411,0.419}
\definecolor{6}{rgb}{0.980,0.54,0.45}
\definecolor{5}{rgb}{0.988,0.666,0.4705}
\definecolor{4}{rgb}{0.990,0.796,0.494}
\definecolor{3}{rgb}{1.00,0.92156,0.5176}
\definecolor{2}{rgb}{0.6941,0.8314,0.498}
\definecolor{1}{rgb}{0.388,0.745,0.4823}

\newcommand{\first}[1]{%
  \begingroup
  \setlength{\fboxsep}{2pt}%
  \colorbox{1}{#1}%
  \endgroup
}
\newcommand{\second}[1]{%
  \begingroup
  \setlength{\fboxsep}{2pt}%
  \colorbox{2}{#1}%
  \endgroup
}
\newcommand{\third}[1]{%
  \begingroup
  \setlength{\fboxsep}{2pt}%
  \colorbox{3}{#1}%
  \endgroup
}
\newcommand{\fourth}[1]{%
  \begingroup
  \setlength{\fboxsep}{2pt}%
  \colorbox{4}{#1}%
  \endgroup
}
\newcommand{\fifth}[1]{%
  \begingroup
  \setlength{\fboxsep}{2pt}%
  \colorbox{5}{#1}%
  \endgroup
}
\newcommand{\sixth}[1]{%
  \begingroup
  \setlength{\fboxsep}{2pt}%
  \colorbox{6}{#1}%
  \endgroup
}
\newcommand{\seventh}[1]{%
  \begingroup
  \setlength{\fboxsep}{2pt}%
  \colorbox{7}{#1}%
  \endgroup
}
\newcommand{\eigth}[1]{%
  \begingroup
  \setlength{\fboxsep}{2pt}%
  \colorbox{8}{#1}%
  \endgroup
}

\newcommand{\eg}{\textit{e}.\textit{g}.}
\newcommand{\ie}{\textit{i}.\textit{e}.}

\newcommand{\Th}{\textsuperscript{th}}

\graphicspath{{Images/}}

\begin{document}

\title{Finding Non-Uniform Quantization Schemes
using Multi-Task Gaussian Processes} 

%
\author{\IEEEauthorblockN{Marcelo Gennari do Nascimento, Theo W. Costain, Victor Adrian Prisacariu}\\
%
%
\IEEEauthorblockA{Active Vision Lab, University of Oxford, UK\\
\tt{\{marcelo,costain,victor\}@robots.ox.ac.uk}\\
\url{https://code.active.vision}}}
\maketitle

\begin{abstract}
We propose a novel method for neural network quantization that casts the neural architecture search problem as one of hyperparameter search to find non-uniform bit distributions throughout the layers of a CNN. We perform the search assuming a Multi-Task Gaussian Processes prior, which splits the problem to multiple tasks, each corresponding to different number of training epochs, and explore the space by sampling those configurations that yield maximum information. We then show that with significantly lower precision in the last layers we achieve a minimal loss of accuracy with appreciable memory savings. We test our findings on the CIFAR10 and ImageNet datasets using the VGG, ResNet and GoogLeNet architectures.
\end{abstract}

\section{Introduction}

The strategy of quantizing neural networks to achieve fast inference has been a popular method of deploying neural networks in compute constrained environments. Its benefits include significant memory savings, improved computational speed, and a decreased cost in the energy needed per inference. Many methods have used this family of strategies, quantizing down to anywhere between 8-bits and 2-bits, with little loss in accuracy\cite{hubara2017quantized,zhang2018lq}.
It also bears noting that in most of these methods, after quantizing to very low precisions (1 to 5 bits), retraining is necessary to recover accuracy.

Recently, even though the quantization algorithms have significantly improved, they have almost exclusively \emph{implicitly} assumed that the best strategy is to quantize all the layers uniformly with the same precision.
However, there are two main reasons to believe otherwise: i) we argue that as it has been interpreted \cite{olah2017feature} that different layers extract different levels of features, it follows that different layers might require different levels of precision; ii) the idea of quantization as an approximation to the floating point (FP) version of the network suggests that lower error in the early layers reduces the propagation of errors down the whole network, minimizing any drop in accuracy.
We believe that as important as having a good quantization strategy, is to also have a good strategy for the distribution of bits through the network, thereby eliminating any redundant bits.
The goal is then to find a configuration in a search space that uses the least amount of bits and achieves the highest accuracy per bit used.

\begin{figure}[ht]
\centering
\includegraphics[width=0.9\linewidth]{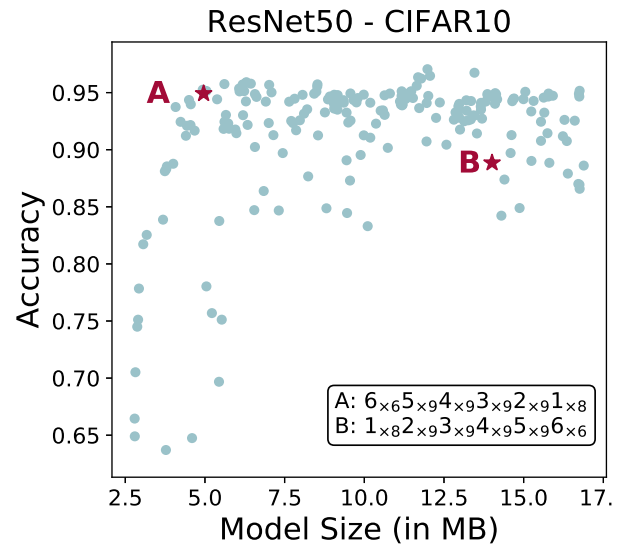}
\caption{Gaussian Process prediction for bit distribution in memory vs accuracy plot}
\label{fig:teaser}
\end{figure}


We cast this Neural Architecture Search (NAS) problem into the framework of hyperparameter search, since the bit-width of each layer should ideally be found automatically.
As with many NAS approaches, measuring the accuracy of a single configuration can take a considerable amount of time.
To mitigate this issue, we propose a two stage approach.
First, we map the full search space into a lower dimensional counterpart through a parameterised constraint function, and second, we use a Multi-task Gaussian Process to predict the accuracy at a higher epoch number from lower epoch numbers.
This approach allows us to reduce both the complexity of the search space as well as the time required to determine the accuracy of a given configuration.
Finally, as our Gaussian Process based approach is suitable for probabalistic inference, we use Bayesian Optimisation (BO) to explore and search the hyperparameter space of variable bit-size configurations.




For the quantization of the network, we use the DSConv method~\cite{Nascimento_2019_ICCV}.
It achieves high accuracy without significant retraining, meaning the number of epochs needed for full training, and implicitly, the requirement for prediction power, is minimised.


To summarise, our main contributions are as follows:
\begin{enumerate}
\item we cast NAS as hyperparameter search, which we apply to the problem of variable bit-size quantization;
\item we reduce the time needed to measure the accuracy of a proposed bit configuration considerably by using multi-task GPs to infer future accuracy from current estimates;
\item we demonstrate performance across a broad range of configurations, described by Bezier curves and Chebyshev series.
\end{enumerate}

The next sections are as follows: Section \ref{sec:related_work} shows previous work on quantization and hyperparameter search. Section \ref{sec:method} elaborates on the methodology used for search, including the constraint, exploration, and sampling procedures. Section \ref{sec:results} shows the results achieved on the CIFAR10  and ImageNet datasets using the networks listed above. Section \ref{sec:conclusion} draws a conclusion and considers insights from the paper.

\section{Related Work}
\label{sec:related_work}

\paragraph{Neural Architecture / Hyperparameter Search} One can consider finding bit distributions as a form of model selection \cite{Rasmussen:2005:GPM:1162254}, given its complexity and the limit on the parameters that it accepts as a solution.
Previous methods have predominantly used Reinforcement Learning (RL) and Evolutionary Algorithms (EA) to model search, which is referred to in the literature as Neural Architecture Search.
Examples include NASNet \cite{zoph2018learning}, MNasNet \cite{tan2019mnasnet}, ReLeq-Net \cite{elthakeb2018releq}, HAQ \cite{wang2019haq}, among others \cite{baker2016designing,zhong2018practical} for RL and \cite{zoph2016neural,kitano1990designing,liu2017hierarchical,stanley2002evolving} for EA.
Our work overlaps with these papers only on the goal of finding an optimal strategy given a search space.

ReLeQ-Net and HAQ, to the best of our knowledge, are the only methods whose aims are to find the optimal bit distribution through different layers of a network, and are therefore the papers that overlap the most with our work.
It is notable that both of them use an RL based approach to search for optimal bit distributions. However, HAQ is more focused on hardware specific optimization, whereas both ours and ReLeQ-Net's methods attempt to be Hardware-Agnostic.
Recently some work involving Bayesian Optimization (BO) for model architecture selection has been carried out, with systems such as NASBOT \cite{k2018neural}.
One of the reasons why BO has not been used for model selection has to do with how unclear it is to find a measure of ``distance" between two models, which is the main problem that was addressed by NASBOT.

Alternatively, one can see determining bit distribution as finding hyperparameters to be tuned given a model, \ie\ not different from finding the optimal learning rates or weight decays.
Historically, this has been tackled by BO techniques.
In neural networks specifically, this was popularized after the work of \cite{snoek2012practical}, and followed by others \cite{Thornton:2013:ACS:2487575.2487629,bergstra2013making,NIPS2015_5872,snoek2015scalable}.
As a result BO can be considered a natural method for searching for optimum bit distribution configurations.

\paragraph{Quantization} Quantization strategies can be either trained from scratch or derived from a pretrained network.
The methods of \cite{zhang2018lq,hubara2017quantized,Cai_2017,zhou2016dorefa} initialize their networks from scratch.
This ensures that there is no initial bias on the values of the parameters, and they can achieve the minimum difference in accuracy when extremely low bit values are used (1-bit / 2-bits) - a notable exception being DoReFa-Net \cite{zhou2016dorefa}, which reportedly had slightly better results when quantizing the network starting from a pretrained network.
The methods of \cite{han2015deep,NIPS2017_6638,Nascimento_2019_ICCV,vanhoucke2011improving} quantize the network starting from a pretrained network.
These methods start with a bias on the values of the parameters, which can limit how much they recover from the lost accuracy.
A benefit of these methods though is that they can be quickly fine-tuned over a few epochs re-achieving state-of-the-art results.
These methods are more interesting to us because of their quick deployment cycle.
It is worth noting that all of these methods use a uniform distribution of precision, meaning that all layers are quantized to the same number of bits.


\section{Method}
\label{sec:method}

Our method consists of three parts: constraining, exploring, and sampling the search space. We first constrain the search space by assuming that the bit in the next layer somewhat depends on the bit used in the current layer.
We do this by drawing bit distributions from a low-degree Polynomial (in the experiments we use a $2^{nd}$ degree Bezier curve and a $4^{th}$ order Chebyshev series). Given a drawn distribution, we quantize the network using the DSConv method. We explore the space by placing a Gaussian prior over the polynomial parameters, and sampling / retraining a set of hyperparameters that gives the most information about the final payoff function. After exploring, we rank the configurations based on sampling the GP for accuracy, and choose the ones that are the most appropriate for our end-use. Each of these phases will be explained further in this section.


\begin{figure*}[t]
    \centering
    \includegraphics[width=0.85\linewidth]{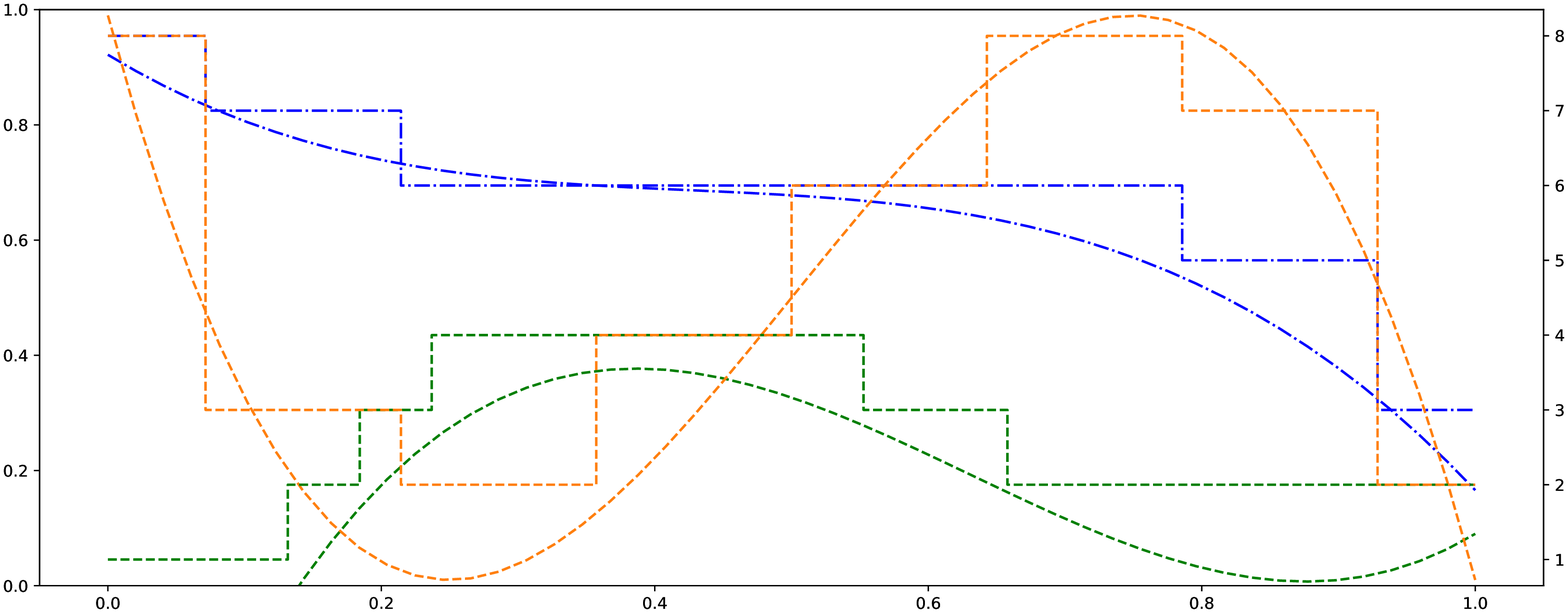}
    \caption{Three examples of modified Chebyshev functions and their clamped versions. The continuous lines represent the values of the modified Chebyshev functions as function of the layer, which is then converted into bitwidths whose values are represented on the right hand axis. This is for two 8 layer VGG11s and a 20 layer ResNet corresponding to configurations 1)blue: 8,7,6,6,6,6,5,3 2) orange: 8,3,2,4,6,8,7,2 and 3) green: 1,1,1,2,3,4,4,4,4,4,4,3,3,2,2,2,2,2,2,2}
    \label{fig:cheb}
\end{figure*}

\subsection{Constraining the Space}
When trying to find the bits, from 1-8, for each layer, the search space will have size of $8^n$, where $n$ is the number of layers of the network. For a CNN of 50 layers, the search space will be $2^{150} \approx 10^{45}$, which is a similar search space to a game of chess ($\approx 10^{50}$). Algorithms have been developed that consistently beat chess grandmasters, however the time for one episode of chess is considerably less than an episode to train and measure the accuracy of a neural network. Therefore, exhaustively searching in this space is prohibitive.

Our method for constraining the search space relies on the use of parameterised functions.
We model a function of degree $n$ with a few hyperparameters, which describe the search space.
From this function, we pick the bit distribution such that it follows the function's curve. In this way, a bit configuration of any layered size network can be sampled from a few hyperparameters alone. 


We use two parameterised functions to illustrate our solution: 
\begin{itemize}
    \item We define the Bezier function $\textbf{B}(x; \mathbf{w}) = \mathbf{w}^T \phi(x)$ for $x \in \mathbb{R}, \mathbf{w} \in \mathbb{R}^d,\ 0 \leq x \leq 1,\ 0 \leq w_i \leq 1\ \forall i\in\{i:i\in\mathbb{N}^+,i<=d\}$, where $d$ is the degree of the polynomial. The vector $\phi(x)$ is the feature vector of the Bezier curve \ie\ for Linear Bezier $\phi(x) = [1-x,\ x]^T$, for Quadratic Bezier $\phi(x) = [(1-x)^2,\ 2(1-x)x,\ x^2]^T$, \textit{etc}.
    \item We define the modified Chebysehv function $\textbf{T}(x; \mathbf{w}) = \frac{(\mathbf{w}-0.5)^T \phi_d(x) +1}{2}$ for $x \in \mathbb{R}, \mathbf{w} \in \mathbb{R}^d, \ -1 \leq x \leq 1, \ 0 \leq w_i \leq 1\ \forall i\in\{i:i\in\mathbb{N}^+,i<=d\}$, where $d$ is the degree of the polynomial. The vector $\phi_d(x)$ is defined as
    $ [ T_0(x), T_1(x), T_2(x), \dots\\, T_{d-1}(x) ]^T$ where $T_0(x) = 1, T_1(x) = x,\ \text{and}\   T_{n+1}(x) = 2xT_{n}(x)-T_{n-1}(x)$
\end{itemize}
The constraint function, $g(t)$ then is a clamped and rounded version of the polynomial function chosen, $p(t)$, such that the bits, $b$,  for each layer generated are between 1 and 8, and $b_i \in \mathbb{N}$. We can define then $g(t) = \lfloor \text{CLAMP}(8p(t)+1), 1, 8) \rceil$, where $\lfloor \cdot \rceil$ is the rounding function, and $\text{CLAMP}(f, a, b) = \min(\max(f, a), b)$. 
 
Fig.~\ref{fig:cheb} shows an example of a Chebyshev function and its clamped version. The y-axis in the left indicate the value of the Chebyshev function for different values of $x$. This is then clamped, rounded, and scaled such that it transforms into a discontinuous line that represents the bit chosen for each layer of a CNN. The bit value is indicated in the y-axis in the right.

By constraining the search space in this way, the minimization problem then shifts as follows:

\begin{equation*}
\begin{aligned}[t]
\quad & \textbf{\text{Na{\"i}ve Approach}}\\
\min_{\textbf{b}} \quad & \mathcal{L}(t; \textbf{b}), \textbf{b}\in\mathbb{N}^n\\
\textrm{s.t.} \quad & 1 \leq b_i \leq 8, \ \forall i \in \{1, n\}\\
\\
\end{aligned}
\end{equation*}
\begin{equation*}
\begin{aligned}[t]
\quad & \ \ \ \ \textbf{\text{Our Approach}} \\
\min_{\textbf{w}} \quad & \mathcal{L}(t; \textbf{w}), \textbf{w}\in\mathbb{R}^d\\
\textrm{s.t.} \quad & b_i = g(\tfrac{i }{n}),\ \forall i \in \{1, n\} \\
\quad & 0 \leq w_j \leq 1, \ \forall j  \in \{1, d\}\\
\end{aligned}
\end{equation*}

where $\mathcal{L}$ is the loss function (to be introduced in Section \ref{subsec:decision_proc}).

The search then reduces to finding the parameters of the polynomial basis $\textbf{w}$, which consequently define the bit distributions throughout the layers. The search space is then continuous and compatible with GPs, and significantly reduced to only $d$ dimensions. Using this parameterisation, we are able to easily define a distance metric between configurations to be used when calculating the kernel function and predictive distribution from our Gaussian Process. So, with this setup, the search space can be sufficiently explored in a timely manner.

\begin{figure*}[tb]
\begin{center}
\includegraphics[width=\linewidth]{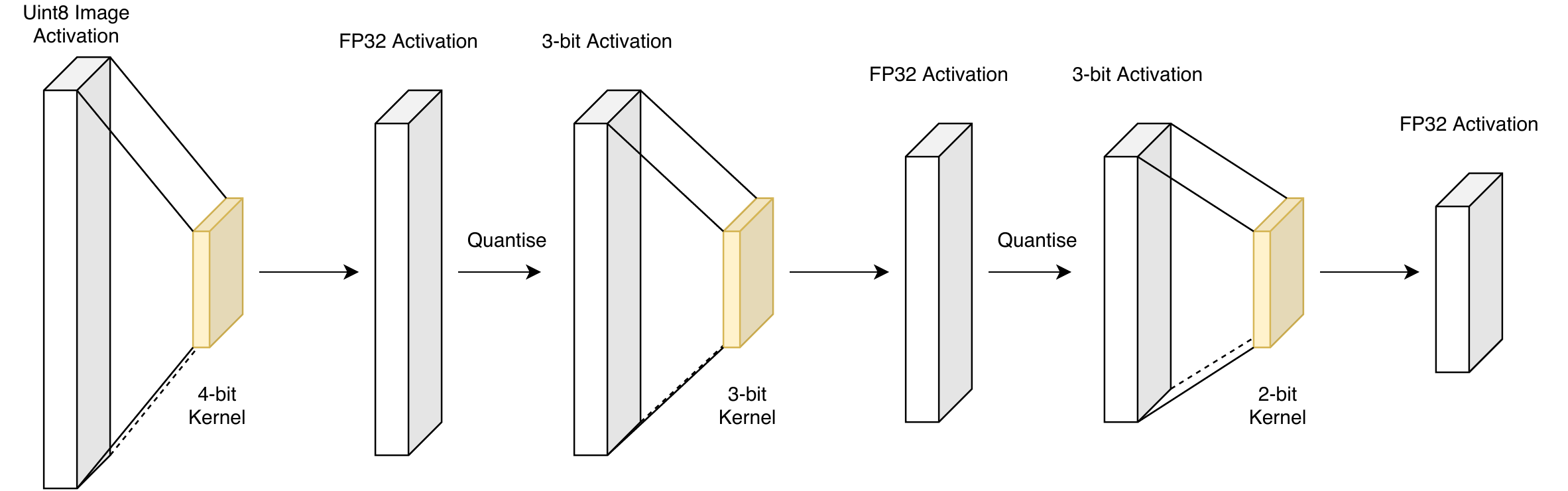}
\end{center}
\caption{Quantization given variable bit-widths. Notice that the input is the image, which is a \texttt{uint8} tensor (normalization can be dumped into a KDS tensor \cite{Nascimento_2019_ICCV}), so it is not quantized. The quantization of activations is done before the convolution such that the convolution can be done using the same precision.}
\label{fig:quantization_image}
\end{figure*}

\paragraph{Quantization Strategy} As mentioned previously, the method used for quantizing the CNN is DSConv. This choice was made because our aim is to minimize time taken during training, and DSConv has consistently shown good accuracy properties in models, even when they are not retrained. This can also be seen as a time-constraint in the search space, such that minimal training time is needed to achieve meaningful accuracy estimations.

In this method, both the activations and the weights are quantized, such that fast inference is possible. Each of the weight tensors are quantized into two tensors, the Variable Quantized Kernel (VQK) and the Kernel Distribution Shift (KDS). Each of them are divided into blocks of size $B=32$ depthwise. The VQK stores integer values in 2s complement, such that $w_q \in \mathbb{Z}, b \in \mathbb{N} \ \text{s.t.} \ -2^{b-1} \leq w_q \leq 2^{b-1} -1$, where $w_q$ are the weight values and $b$ is the number of bits in that layer. This has the same size as the FP32 weight tensor, and its values are found by simply scaling each of the block from the original FP32 weight tensor by $\frac{2^{b-1}}{\text{MAX}(w)}$, and then flooring and cropping to range. The KDS is a tensor $B$ times shallower, that holds FP32/16 scaling values, each which corresponds to a block of the VQK. Their values are calculated by simply minimizing the L2 norm of each block with respect to the original corresponding block: $\xi = \frac{\sum_{i=0}^{B-1}w_iw_{qi}}{\sum_{i=0}^{B-1} w_{qi}^2}$. The idea is that at the end of the quantization process, the KDS multiplied by the VQK - correspondingly with each block - will be as similar as possible to the original FP32 weight tensor. The activation tensor is quantized similarly, but using Block Floating Point (BFP) format in each of the blocks instead.

In order to take advantage of the low bit multiplication speed, the activation tensor and the weight tensor need to have the same precision. Figure \ref{fig:quantization_image} shows how this is done. The activation tensor prior to a convolution layer is set to be quantized to the same bit precision as that layer. The first convolution is not quantized since the input image is already in \texttt{uint8} format. In this way, a quantization distribution strategy can be fully defined by providing the precision on each of the layers. Also note that we quantize only the convolutional layers. The Fully Connected layers are all left in the original FP32 precision for training.

This quantization strategy has shown good properties with little to no retraining at all. Since our goal is to show that uneven distribution of precision through the layers of a network is a better strategy for quantization, we will use the fact that DSConv needs little to no retraining in order to accelerate our search algorithm. Whereas multiple other algorithms need many epochs to at least achieve optimum accuracy, DSConv will only need a few iterations before meaningful results are achieved.

\begin{figure*}[t]
\begin{center}
\includegraphics[width=\linewidth]{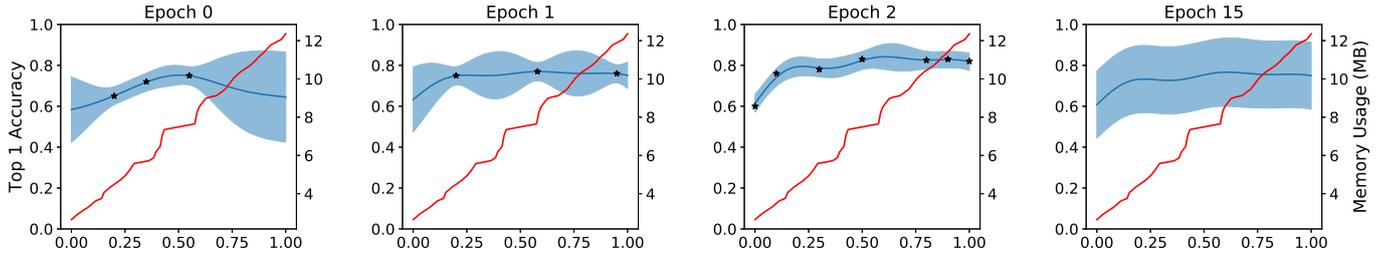}
\end{center}
\caption{Multi-Task Gaussian Process for inferring accuracy of quantized network. The quantization function is a Bezier Linear with the first parameter set to 0.5 \ie\ $g(t; w_1) = 0.5+t(w_1 - 0.5)$. For all figures, the x-axis is the value of $w_0$, the left y-axis is the accuracy on CIFAR10 of a toy CNN with 10 layers. The right y-axis (red line) shows the model size for a given value of $w_0$. The epoch correspondences for each task is [0, 1, 2, 15] respectively. After this exploration phase, the decision procedure is run on the predictive distribution of Task 4.}
\label{fig:1d_gp}
\end{figure*}

\subsection{Exploring the Space}
\label{subsec:exploration}
Next, we need a way of exploring the space in order to learn the accuracy of the network given a limited set of $\textbf{w}$ points. We propose a Multi-Task Gaussian Process prior in the neural network, such that each task corresponds to the estimation of the accuracy of the quantized neural networks given the parameter $\textbf{w}$ after a certain number of epochs \eg\ task 1 corresponds to 0 epochs, task 2 to 1 epoch, task 3 to 2 epochs, task 4 to 15 epochs. Let there be $m$ tasks, and a prior on $f_l$, $l \in \{0, m\}$, such that $f_l \sim  \mathcal{GP}(\mu(t), k(t, t'))$. We also place a probability distribution $\mathcal{P}(f_0, f_1, ... f_m)$ over different tasks. Let $y_{(t, l)} = f_l(t) + \epsilon(t) $ be the observation at hyperparameter value $t$ for task $l$, and let $\epsilon(t)  \sim \mathcal{N}(0, \sigma^2; t)$ be the observation noise, which is normally distributed. This defines independent Gaussian Likelihoods $y_{(t, l)} \sim \mathcal{N}(f_l, \sigma^2; t)$. From this model, observations $\mathbf{y}$ are drawn, such that $\mathbf{y} = (y_{11}, ..., y_{s1} ..., y_{12}, ..., y_{s2}, ..., y_{1m}, ..., y_{sm})$, where $y_{il}$ is the $i^{th}$ observation of the $l^{th}$ task.

We used the The Intrinsic Correlation Model (ICM) of \cite{chai2010multi} and \cite{bonilla2008multi} for kernel calculation (in our experiments we made use of the squared exponential kernel). We can then define the mean and the correlation between tasks as:
\begin{equation}
    \begin{aligned}
        \langle f_l(x)\rangle &= \mu_l (x)
        \\
        \mathbb{C}(f_l(x), f_{l^\prime}(x^\prime)) &= k^f(l, l^\prime)k^x(x, x^\prime)
    \end{aligned}
\end{equation}
where $k^f$ and $k^x$ are positive semi definite functions, corresponding to the correlation between functions and the correlation between inputs respectively. From this it follows that the covariance is $K =K^f \otimes K^x$, where $\otimes$ is the Kronecker product, $K^f$ is the matrix of correlations between the functions and $K^x$ is the matrix of correlations between the inputs. For a new set of data points $x_*$, the mean prediction can then be calculated using the normal formula for the predictive distribution:
\begin{equation}
    \begin{aligned}
        \overline{\mathbf{f}}(x_*) &= \mu_l(x_*) + (K^{x_*})^T \Sigma^{-1}\mathbf{y}
        \\
        \Sigma &= K + D \otimes I
    \end{aligned}
\end{equation}
where $D$ is an $m \times m$ diagonal matrix where the $(l, l)^{th}$ term is $\sigma_l^2$.

Figure \ref{fig:1d_gp} shows an example of the Multi-Task setting with a 1D Bezier Curve for ease of visualization. Each plot shows the predictive mean and variance for each epoch after 14 data points have been collected, using the exploration algorithm explained in section \ref{subsec:exploration}, \eg\ Task-1 was set to 0 epochs (so straight quantizing from FP32 model); Task-2 was set to 1 epoch; Task-3 to 2 and Task-4 to 15. The idea is to predict what is the distribution of the last task given inputs in earlier tasks.

\paragraph{Exploration Phase}
In order to make decisions on what parameters to choose, we need to explore the space to predict the accuracy of the last task. The exploration phase for the multitask Gaussian Process follows the Low-Fidelity Search from \cite{song2018general}. The idea is to find the values of $x, l$ such that it gives us maximal information $\mathbb{I} (y_{(x, l)}; f_m |\  \textbf{y}) = \mathbb{H}(y_{(x, l)} \ |\ \textbf{y}) - \mathbb{H}(y_{(x, l)} \ | \ \textbf{y}, f_m)$, where $\textbf{y}$ is the observation history, and $(x, l)$ is the action to be performed. It is important to weight the information by a measure of the cost that it takes to perform that operation. So the exploration procedure chooses $x, l$ that maximizes $\mathbb{I} (y_{(x, l)}; f_m |\  \textbf{y})$ per unit cost. This means that the parameter that has the most information about the payoff function will be picked.

Depending on the dataset and model chosen, the user can favour exploration on one fidelity over the other by decreasing the cost $\lambda$ of running that particular task.
Additionally, we set up a budget on the amount of time in unit cost or number of architectures that we are willing to explore. The Exploration Phase finishes when the Budget has been fully used. After this is finished, the user can run their preferred method of ranking configurations using the posterior of the trained GP.

\subsection{Sampling the Space}
\label{subsec:decision_proc}

The na{\"i}ve goal is to find the highest accuracy per bit possible, which corresponds to finding the minimum of the loss function $\mathcal{L}(t; \textbf{w}) = - \frac{y_{(t, m)}}{\sum_{i=1}^{n} b_i}$. However, there is a trade-off that must be considered. A model, \eg\ ResNet20, using a total of 40 bits and achieving
80\% accuracy (ratio of 2\%/bit) is arguably worse than a model that uses 43 bits and achieves 85\% accuracy (ratio of 1.97\%/bit). The goal is instead to find a decision procedure that takes into account the regret of not using more bits based on a set of constraints. This relationship should be linear instead of inversely proportional. A better strategy instead is to assume that using 4-bits for all layers is the best that can be done when quantizing without losing accuracy. Each bit used less than this should be a reward, and each bit used more than this should be a penalty, this is added (or subtracted) to the accuracy to get an ``effective accuracy". We then define the effective accuracy as $\mathcal{E}(a, \textbf{b}, n) = a - \frac{\sum_{i=1}^{n} b_i - 4n}{k}$, where $a$ is the accuracy of the original network, and $k$ is a constant of penalty per bit. Therefore, for $k=100$ each bit used in addition to the average of 4-bits incurs a penalty of 1\% in the effective accuracy. The reverse incurs a reward of 1\% in the effective accuracy. The decision procedure becomes then to minimize the negative effective accuracy, $\mathcal{L} = - \mathcal{E}(a, \textbf{b}, n)$. Once we have enough information about the GPs, we can rank configurations based on their loss in order to pick the most relevant for us.


\section{Experiments and Results}
\label{sec:results}
We tested our method in a variety of configurations, using versions of the original VGG, ResNet, and GoogLeNet models, altered in order to take CIFAR10, and ImageNet32 as input. For training CIFAR10 and ImageNet32, we used data augmentation by cropping 32x32 image of the 4-pixel padded original. We used a Stochastic Gradient Descent optimiser with momentum of 0.9, and weight decay of $5\times10^{-5}$. The learning rate started equal to $10^{-1}$, and was divided by 10 after 150 and 250 epochs.  

We ran the exploration procedure on $\sim 65$ configurations for each network using the multi-task algorithm outlined above. From these configurations, we could then use the mean of the gaussian to draw estimates of the accuracy of many different quantization schemes. Using the decision outlined above, we sorted the results by either accuracy, memory, or computational complexity, and selected the points of interest for better visualization and intuition of what the general trend of the found configurations are.

\begin{table*}[htbp]
\caption{Results for many configurations on CIFAR10. VGG16 and VGG19 correspond to the architectures introduced in~\cite{simonyan2014very}. ResNet18 is the architecture from \cite{he2016deep}, and the GoogLeNet architecture is from \cite{szegedy2015going}. The Configuration refers to the bit value for each layer of a given model, from earlier layers in the left to later layers in the right. They are color coded for clarity: red for higher bits and green for lower bits. It is important to note that we quantize only the convolutional layers, which means that VGG16 has 13 values, VGG19 has 16 values, ResNet18 has 20 values, and GoogLeNet has 64. Because of its size, the GoogLeNet values were represented by a subscript indicating the number of times that a given bit-width is used. The column ``Delta" refers to the difference between the GP estimation of the Top1 accuracy and the actual mean Top1 accuracy ($n=10$) after properly retraining that particular configuration.}
\label{tab:ResultsLinear}
\centering
\resizebox{0.99\linewidth}{!}{
\begin{tabular}{cccccccc}\toprule
\textbf{CNN} & \textbf{\begin{tabular}[c]{@{}c@{}}Configuration\\ (bits per layer)\end{tabular}} & \textbf{\begin{tabular}[c]{@{}c@{}}Top 1 Estimate\\ from GP\end{tabular}} & \textbf{\begin{tabular}[c]{@{}c@{}}Mean\\ Top1\end{tabular}} & \textbf{Std} & \textbf{Delta} & \textbf{\# Bits} & \textbf{\begin{tabular}[c]{@{}c@{}}Memory\\ (in MB)\end{tabular}} \\ \midrule 
\multirow{7}{*}{{VGG16}} & 32-bit Floating Point & -  & 93.7\%  & -   & - & -  & 58.8  \\
&\sixth{6}\fifth{555}\fourth{44}\third{333}\second{22}\first{11}& (95.5\%)& \textbf{93.7\%}  & 0.2\% & -1.8\% & 50 & \textbf{4.84} \\
& \first{11}\second{22}\third{333}\fourth{44}\fifth{555}\sixth{6} & (91.3\%) & \textbf{87.7\%}& 0.2\% &{-3.6\%}& 50 & \textbf{9.50} \\
& \seventh{7}\sixth{66}\fifth{55}\fourth{44}\third{33}\second{222}\first{1} & (92.1\%)  & 93.7\%  & 0.1\% & {1.6\%}& 50 & 5.26 \\
& \first{1}\second{222}\third{33}\fourth{44}\fifth{55}\sixth{66}\seventh{7} & (90.1\%)  & 91.5\%  & 0.4\% & {1.4\%} & 50 & 10.75 \\
& {\fourth{4444444444444}}                                                  & (93.3\%) & 93.8\%  & 0.1\%  & {0.5\%}& 52 & 8.28 \\
& {\third{3333333333333}}                                                   & (92.9\%)  & 93.5\%  & 0.2\% & {0.6\%} & 39 & 6.44 \\ \addlinespace[2em]
\multirow{7}{*}{{VGG19}}    & 32-bit Floating Point & - & 93.9\% & -  &- &-  & 80.1 \\
& \sixth{6}\fifth{555}\fourth{4444}\third{333}\second{222}\first{11} & (94.4\%) & \textbf{93.7\%}  &  0.1\% & {-0.7\%} & 54  & \textbf{6.95} \\
& \first{11}\second{222}\third{333}\fourth{4444}\fifth{555}\sixth{6} & (91.6\%)& \textbf{89.6\%}&0.4\%& {-2.0\%} & 54& \textbf{12.04}\\
& \fifth{5}\fourth{4444}\third{33333}\second{2222}\first{11}& (93.9\%)&93.5\%&0.1\%& {-0.4\%} & 46& 6.14\\
& \first{11}\second{2222}\third{33333}\fourth{4444}\fifth{5}& (90.3\%)&88.4\%&1.2\%& {-0.9\%}& 46& 10.05\\
& \third{3333333333333333}& (92.9\%)&93.4\%&0.2\%& {0.5\%} & 48& 8.76\\
& \second{2222222222222222}& (92.1\%)&92.2\%&0.2\%& {0.1\%} &32& 6.25\\ \addlinespace[2em]
\multirow{7}{*}{{ResNet18}} & 32-bit Floating Point& -& 95.4\%& -&- &-&  44.6\\
& \sixth{666}\fifth{5555}\fourth{4444}\third{3333}\second{2222}\first{1}   & (96.3\%)&\textbf{95.4\%}&0.1\%& {-0.9\%}&75& \textbf{3.72}\\
& \first{1}\second{2222}\third{3333}\fourth{4444}\fifth{5555}\sixth{666}   & (95.9\%)&\textbf{92.9\%}&0.3\%   &{-3.0\%}     & 75               & \textbf{8.00}\\
& \fourth{44444444}\third{3333333333}\second{22}   & (95.3\%)&     95.3\%&0.1\%& 0.0\% & 60& 4.34\\
& \second{22}\third{3333333333}\fourth{44444444}   & (94.5\%)&94.2\%&0.2\%& {-0.3\%} &66& 6.08\\
& \third{33333333333333333333}   & (94.4\%)&     95.0\%&0.1\%& {0.6\%}& 60& 4.90\\
& \second{22222222222222222222}   & (93.1\%)&       93.3\% &0.5\%& {0.2\%}& 40& 3.49\\\addlinespace[2em]
\multirow{7}{*}{{GoogLeNet}} & 32-bit Floating Point& -& 95.5\%& -&- &-&  24.32\\
& \fourth{4}{\scriptsize$\times$21} \third{3}{\scriptsize$\times$27} \second{2}{\scriptsize$\times$16}   & (94.7\%)&     \textbf{95.3\%}&     0.1\%& 0.6\% & 207 & \textbf{2.35}\\
& \second{2} {\scriptsize$\times$16} \third{3} {\scriptsize$\times$27} \fourth{4} {\scriptsize$\times$21}   & (94.6\%)&\textbf{94.2\%}&0.1\%& {-0.4\%} &207& \textbf{2.98}\\
& \sixth{6}{\scriptsize$\times$8} \fifth{5}{\scriptsize$\times$13} \fourth{4}{\scriptsize$\times$12} \third{3}{\scriptsize$\times$13} \second{2}{\scriptsize$\times$13} \first{1}{\scriptsize$\times$5}   & (95.8\%)&95.3\%&0.1\%& {-0.5\%}&231& 2.65\\
& \first{1}{\scriptsize$\times$5} \second{2}{\scriptsize$\times$13} \third{3}{\scriptsize$\times$13} \fourth{4}{\scriptsize$\times$12} \fifth{5}{\scriptsize$\times$14} \sixth{6}{\scriptsize$\times$8}   & (94.7\%)&90.5\%&       0.2\%   &{-4.2\%}     & 231               & 3.73\\
& \third{3}{\scriptsize$\times$64}   & (94.7\%)&     95.1\%&     0.1\%& {0.4\%}& 192 & 2.68\\
& \second{2}{\scriptsize$\times$64}   & (93.4\%)&       93.5\%        &      0.2\%& {0.1\%}& 127 & 1.92\\
\bottomrule
\end{tabular}
}
\end{table*}

\paragraph{Results on Accuracy using the CIFAR10 Dataset}
Results on CIFAR10 and ablation tests are displayed in Table \ref{tab:ResultsLinear}. The configurations are color coded for clarity, with red representing higher bit counts and green representing lower bit counts. These configurations were selected based on the decision procedure outlined above, using the Bezier Linear polynomials.

For comparison, we show 6 configurations of each network: the first and third configurations were picked by the decision procedure outlined above; the second and fourth configurations are simply the inverse order of the first and third configurations; the fourth and fifth rows use the traditional uniform distribution of bits for a fair comparison.

It is important to note that the decision to pick these configurations are based on the estimate of the GP rather than on the actual Top-1 and Top-5 results. In order to compare fairly, we also included the Top-1 and Top-5 scores after properly training each of them for an additional 30 epochs using the same hyperparameters and optimiser that were used the train the FP32 version of each of these networks. We have also included a delta column which shows the difference between the Top-1 estimate from the GP and the Top-1 after fine-tuning the network. It is remarkable that most of the error in estimation is within 1\%, which shows how the GP was able to generalize and interpolate properly as expected.

It can be seen that in general, using more bits in earlier layers yields more accurate, and lighter configurations. The higher accuracy can be explained numerically, since higher bits are used in earlier layers, the error propagation through the network is smaller. The lower memory usage is due to the fact that later layers have a higher number of channels, and therefore using lower precision in those layers yield a massive difference in memory need. For VGG16, the first configuration is both lighter, faster, and more accurate than using 3-bits for all layers. This pattern is repeated for the deeper VGG19 too, where the first configuration yielded superior results to the constant 3-bits for all layers, and also for ResNet18 as well. This ``rule of thumb" is somewhat weaker in the GoogLeNet architecture though, even though there is still a clear correlation.

\paragraph{Results on Accuracy using Chebyshev Series}
In order to test robustness of the method in relation to the choice of prior functions, we chose to use a Chebyshev Series of fourth degree, which has a larger search space than the Bezier Linear model. We have tested the model using the CIFAR10 dataset as well, and the results are shown in Table~\ref{tab:ChebyshevResults}.

As it can be seen, the $4^{th}$ degree introduced more flexibility as to what bit configurations the method is capable of finding. We found that with higher degree of polynomials, the number of architectures to search should also increase. In our experiments, we have searched for $\sim 150$ configurations before finding good results. 
The table shows the expected result that more bits at the beginning compensate for the fewer bits at the end of the network. The ResNet-18 result resembles the configuration found in Table \ref{sec:results}, even though it found a configuration that has more usage of 3-bits, but performs slightly worse. As also expected, when the bit distribution is inverted in the network, it results in both higher memory and lower accuracy.

The same behaviour is found with the VGGs, with the slight difference that as VGG11 is too shallow, it requires more bits to recover the accuracy. VGG16 is considerably deeper, and therefore our algorithm was able to compress it more significantly.

This results shows that our method can be used with a variety of basis. It is worth bearing in mind that the GP processing capability requires the inversion of a matrix, which is proportional to the degree of the polynomial chosen.
Therefore our method will only work in a timely manner when using fewer hyperparameters to describe the function.

\begin{table}[ht]
\caption{Results of method when using Chebyshev Polynomials of 4\Th degree.}
\label{tab:ChebyshevResults}
\resizebox{.99\linewidth}{!}{
\centering
\begin{tabular}{@{}ccccc@{}}
\toprule
Method & Network & Bitwidths & \begin{tabular}[c]{@{}c@{}}Accuracy \\ Loss\end{tabular} & \begin{tabular}[c]{@{}c@{}} Memory \\(as a \% of original)\end{tabular}\\ \midrule
Ours   & ResNet18 & \begin{tabular}[c]{@{}c@{}}\small{\sixth{6}\fourth{4}\third{33}\second{222}\third{33333333}\second{22}}\\\small{\second{22}\third{33333333}\second{222}\third{33}\fourth{4}\sixth{6}}\end{tabular}  & \begin{tabular}[c]{@{}c@{}}-0.6\% \\ -1.2\% \end{tabular} & \begin{tabular}[c]{@{}c@{}} 8.0\% \\ 11.5\%\end{tabular}                                              \\\addlinespace[0.5em]
Ours   & VGG11     & \begin{tabular}[c]{@{}c@{}}\small{\seventh{7}\sixth{666}\seventh{7}\sixth{6}\fifth{5}\fourth{4}}\\\small{\fourth{4}\fifth{5}\sixth{6}\seventh{7}\sixth{666}\seventh{7}}\end{tabular}                                                    & \begin{tabular}[c]{@{}c@{}}-0.1\%\\-0.5\%\end{tabular}    & \begin{tabular}[c]{@{}c@{}}16.8\% \\ 19.7\%\end{tabular}                                           \\ \addlinespace[0.5em]
Ours   & VGG16     & \begin{tabular}[c]{@{}c@{}}\small{\fourth{4}\third{3}\second{222}\third{3333}\second{22}\first{11}}\\\small{\first{11}\second{22}\third{3333}\second{222}\third{3}\fourth{4}}\end{tabular}        & \begin{tabular}[c]{@{}c@{}}-0.8\%\\-2.4\%\end{tabular}  & \begin{tabular}[c]{@{}c@{}}6.3\%\\ 8.3\%\end{tabular} \\ \bottomrule
\end{tabular}
}
\end{table}

\begin{figure*}[htbp]
\begin{center}
\includegraphics[width=.9\linewidth]{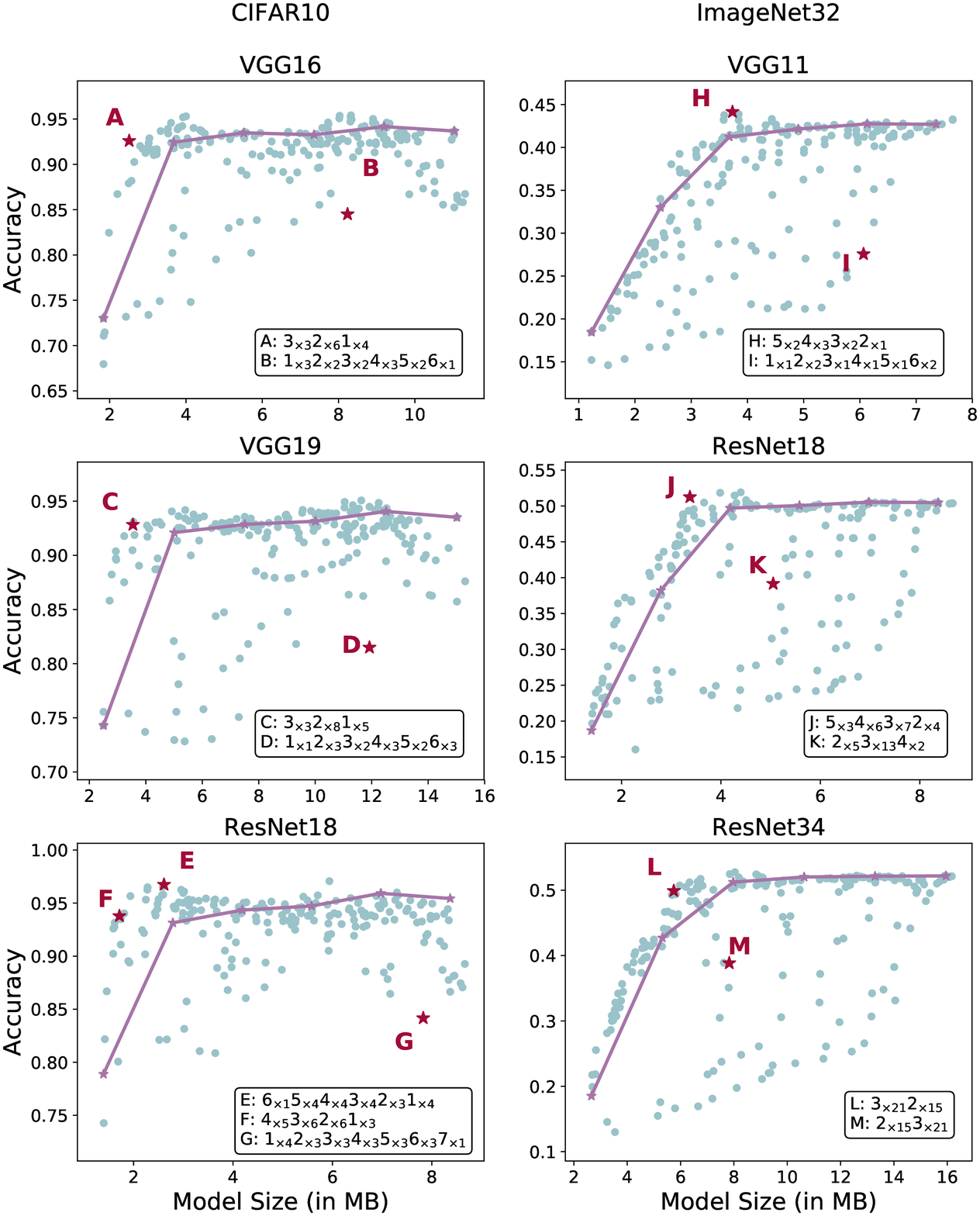}
\end{center}
\caption{Scatter plot of the effect on accuracy versus model size of different bit configurations. The left three plots use the CIFAR10 dataset and the right three plots use the ImageNet32 dataset. Note that this is the plot of the estimate as given by the trained GP, and not the actual accuracy given proper training. The solid line refers to the uniform configurations, starting with all 1s and ending with all 6s. Points A-M highlight different configurations as shown in the text boxes. The string of numbers shown refers to the bit size on each layer of the given network. Note that VGG11 has 8 convolutional layers, and therefore points A and B have only 8 numbers. This applies to VGG16 (13 layers), VGG19 (16 layers), ResNet18 (20 layers), and ResNet34 (36 layers) as well. }

\label{fig:AccuracyByParameter}
\end{figure*}

\paragraph{Results on Network Size}
Figure \ref{fig:AccuracyByParameter} shows the result of the GP-estimated accuracy of different configurations by their model size. The solid purple line links the uniform configurations, starting with all 1s and finishing with all 6s. Therefore, any point that lies above that line is an interesting point, since it gives better accuracy by using the same amount of memory of its uniform counterpart. We have highlighted a number of different interesting configurations with red stars and labelled them from A-M in order to better visualise what each point represent.

As it can be seen in the figure, the choice of bit-usage throughout the network plays an important role in both the accuracy and the memory usage. Even though there is a clear trend that links model size and accuracy, there are a handful of configurations which can perform well on both fronts. It can be seen that, in general, points that are above the purple line are linearly decreasing with bit-usage whereas the ones that are below the purple line are linearly increasing with bit-usage.

The surprising result is that, in the CIFAR10 experiments, even though using uniformly 1-bit for all layers achieves bad results, by just introducing a couple of bits in the first three quarters of the network (such as in points A, C and F), the memory increase is almost negligible, but the accuracy recovery is significant.
Adding bits at the end of the network however, achieves the opposite effect.
It can also be noticed that points A, C, E, and F, achieve better accuracy as the uniformly 2s configuration whilst using 50\% less memory. This is even more evident in point E, in which we used up to 6 bits in the first layers, but still achieved less memory usage due to the usage of 1-bit in the bigger kernels at the end of the network.

In the ImageNet32 experiments, we also see some improvement, albeit less dramatic than the CIFAR10 experiments. The overall massage is still the same, as it can be seen in points H, J and L, for which adding bits in the first layers has achieved good accuracy with small memory increases. It is still noteworthy that even with a dataset as challenging as ImageNet32, due to its substantial decrease of information when compared to the default ImageNet, the GP could find good configurations without needing more datapoints. This shows that this method can be robust to changes in dataset.

\paragraph{Brief Comparison with ReLeQ} One of the other papers that touched in this subject was ReLeQ \cite{elthakeb2018releq}. As explained in the literature review section, they use a reinforcement learning approach to find optimum bit-distributions over the network. Whilst their quantization methodology varies greatly from that used in this paper, it is worth comparing their results to ours. Their results for the CIFAR10 dataset in two of the networks are shown in Table~\ref{tab:comparison}. It can be seen that we achieve similar results for ResNet, though with different mean bits. Since ReLeQ's method does not use the same constraint as our method, it could find more varied solutions. This is a limitation to our method which allows it to find solutions to the network faster whilst using less computational power, but reduces the freedom of choice.

\begin{table*}[htb]
\caption{Comparison of our accuracy results with ReLeQ's method~\cite{elthakeb2018releq} on CIFAR10. The authors in \cite{elthakeb2018releq} did not provide their models size in memory, so we estimated using both our and the original author's quantisation scheme to make a fair comparison.}
\label{tab:comparison}
\centering
\begin{tabular}{@{}cc@{\quad}cccc@{}}
\toprule
Method & Network   & Bitwidths & \begin{tabular}[c]{@{}c@{}}Accuracy \\ Loss\end{tabular} & \multicolumn{2}{c}{\begin{tabular}{cc}\multicolumn{2}{c}{Model size (MB)}\\\midrule
DSConv & WRPNx1 \end{tabular}}\\ \midrule
ReLeQ \cite{elthakeb2018releq} & ResNet-20 & \small{\eigth{8}\second{22}\third{3}\second{222}\third{3}\second{2}\third{333}\second{222}\third{3}\second{2222}\eigth{8}} & 0.12\% & 3.88 & 3.25 \\\addlinespace[0.5em]
Ours   & ResNet-20 & \small{\sixth{666}\fifth{5555}\fourth{4444}\third{3333}\second{2222}\first{1}}  & 0.1\% & 3.54 & 2.91\\\addlinespace[0.5em]
ReLeQ \cite{elthakeb2018releq} & VGG-11    & \eigth{8}\fifth{5}\eigth{8}\fifth{5}\sixth{6666}\eigth{8} & 0.17\% & 6.86 & 6.61\\\addlinespace[0.5em]
Ours   & VGG-11     & \small{\sixth{777}\fifth{666}\fourth{55}}  & 0.14\% & 6.35& 5.42 \\\addlinespace[0.5em]
ReLeQ \cite{elthakeb2018releq} & VGG-16 & \eigth{888}\sixth{6}\eigth{8}\sixth{6}\eigth{8}\sixth{6}\eigth{8}\sixth{6}\eigth{8}\sixth{6}\eigth{8}\sixth{6}\eigth{88} & 0.1\% & 13.32& 12.54\\\addlinespace[0.5em]
Ours   & VGG16     & \small{\sixth{6}\fifth{555}\fourth{44}\third{333}\second{22}\first{11}}   & 0.1\%  & 4.62&3.74 \\ \bottomrule
\end{tabular}
\end{table*}

\paragraph{Results on ImageNet using ResNet} For completeness, we have included some of the results found by our algorithm on the more challenging ImageNet dataset \cite{ILSVRC15}. We decided to use the dataset in order to have results comparable with other methods. This was trained using an Adam Optimizer \cite{kingma2014adam}, with learning rate of $10^{-5}$.

Table~\ref{tab:imagenet} shows the results. As expected, the same pattern of decreasing precision downstream holds across datasets. Comparing these results with the results from DSConv \cite{Nascimento_2019_ICCV}, we can see that a decreasing bit-width throughout the architecture, starting with 6-bits and finishing with 2-bits, which is superior to the ``all 4s'' and ``all 3s''. 

\begin{table}[htb]
\caption{Results of our method using the ImageNet dataset with the ResNet architecture.}
\label{tab:imagenet}
\centering
\resizebox{.99\linewidth}{!}{
\renewcommand{\arraystretch}{1.2}
\begin{tabular}{@{}cc@{\quad}c@{\quad}cc@{}}
\toprule
Method & \# of Layers  & Bitwidths  & \begin{tabular}[c]{@{}c@{}}Acc. \\ Loss\end{tabular} &  \begin{tabular}[c]{@{}c@{}}Size \\ (MB)\end{tabular} \\ \midrule
Ours    & 18 & \sixth{6}{\scriptsize$\times$3} \fifth{5}{\scriptsize$\times$5}  \fourth{4}{\scriptsize$\times$5} \third{3}{\scriptsize$\times$5} \second{2}{\scriptsize$\times$2}           & 0.2\% & \textbf{4.89}\\
DSConv \cite{Nascimento_2019_ICCV} & 18 & 4             & 0.0\% & 5.88\\
DSConv \cite{Nascimento_2019_ICCV} & 18 & 3             & 0.8\% & 4.55\\
Ours    & 50 & \sixth{6}{\scriptsize$\times$6} \fifth{5}{\scriptsize$\times$15}  \fourth{4}{\scriptsize$\times$14} \third{3}{\scriptsize$\times$15} \second{2}{\scriptsize$\times$3}             & 0.6\% & \textbf{11.89}\\
DSConv \cite{Nascimento_2019_ICCV} & 50 & 4             & 0.0\% & 14.54\\
DSConv \cite{Nascimento_2019_ICCV} & 50 & 3             & 0.9\% & 11.74\\
HAQ \cite{wang2019haq}     & 50 & \emph{flexible} & 0.0\% & 12.14\\\bottomrule
\end{tabular}
}
\end{table}

As with ReLeQ's method, HAQ's method has a weaker constraint on bit distribution, which means it would be able to find configurations that our method would not;
 However, even with our very strong constraint, we were still able to find configurations that are competitive in memory requirements to those found by HAQ.
This shows the strength of the conclusion that later layers require lower precision than earlier layers to maintain the same accuracy.

\section{Conclusion}

In this paper, we demonstrate that a uniform distribution over bit-widths throughout a CNN is likely not the most efficient way to quantize a neural network. In order to demonstrate this, we used a Multi-Task Gaussian Process prior over different training epochs, and a Bayesian Optimization exploration procedure based on Information Maximization that estimated the accuracy of different configurations.

We have observed that starting a CNN with higher bit-widths and decreasing precision in later layers yield better accuracy and better memory usage than the traditional uniformly distributed bit-width. This can be interpreted either numerically (as in less error being propagate down the network), or can be interpreted as the functionality of each layer in the network (earlier layers are concerned with feature extraction and later layers are concerned with classification).
\label{sec:conclusion}

\section{Acknowledgements}
This research was supported by Intel and the EPSRC, and we thank our colleagues from the Programmable Solutions Group who greatly assisted in this work.

%
%
\bibliographystyle{splncs04}
\bibliography{egbib}
\end{document}